\documentclass[conference]{IEEEtran}
\usepackage{cite}
\ifCLASSINFOpdf
   \usepackage[pdftex]{graphicx}
\else
\fi
\usepackage[cmex10]{amsmath}

\usepackage{pdflscape}

\usepackage{hyperref}
\usepackage[ justification=centering, font=footnotesize]{caption}
\usepackage[caption=false, font=footnotesize]{subfig}
\usepackage{gensymb}

\begin{document}
\title{Speech-Gesture Mapping and Engagement Evaluation in Human Robot Interaction}

\author{\IEEEauthorblockN{Bishal Ghosh}
\IEEEauthorblockA{Department of Mechanical Engineering\\
Indian Institute of Technology\\
Ropar, India\\
Email: ghosh.bishal@outlook.in}
\and
\IEEEauthorblockN{Abhinav Dhall}
\IEEEauthorblockA{Department of Computer Science\\ and Engineering\\
Indian Institute of Technology\\
Ropar, India\\
Email: dhallabhinav@gmail.com}
\and
\IEEEauthorblockN{Ekta Singla}
\IEEEauthorblockA{Department of Mechanical Engineering\\
Indian Institute of Technology\\
Ropar, India \\
Email: Ekta@iitrpr.ac.in\\
}}


%Telephone: (800) 555--1212, Fax: (888) 555--1212}


\maketitle

\begin{abstract}
%\boldmath
A robot needs contextual awareness, effective speech production and complementing non-verbal gestures for successful communication in society. In this paper, we present our end-to-end system that tries to enhance the effectiveness of non-verbal gestures. For achieving this, we identified prominently used gestures in performances by TED speakers and mapped them to their corresponding speech context and modulated speech based upon the attention of the listener.  The proposed method utilized Convolutional Pose Machine \cite{c4} to detect the human gesture. Dominant gestures of TED speakers were used for learning the gesture-to-speech mapping. The speeches by them were used for training the model. We also evaluated the engagement of the robot with people by conducting a social survey. The effectiveness of the performance was monitored by the robot and it self-improvised its speech pattern on the basis of the attention level of the audience, which was calculated using visual feedback from the camera. The effectiveness of interaction as well as the decisions made during improvisation was further evaluated based on the head-pose detection and interaction survey.
\end{abstract}


% \fi
\IEEEpeerreviewmaketitle

\section{Introduction}
Earlier robots were secluded from interacting with humans fearing that it might harm any human nearby. This general trend is now shifting towards an intermingled society of humans and robots working in synchronization. Many such initiatives which try to dilute the boundary between human and robots have been seen in the past decade. Few of such initiatives are Ashimo\footnote{http://asimo.honda.com/}, NAO\footnote{https://www.softbankrobotics.com/emea/en/robots/nao} and Aibo\footnote{https://aibo.sony.jp/en/}.

\begin{figure}[t]
      \centering
     
      \subfloat[Image showing head pose of attendees as seen by NAO during attention tracking.]
        {
            \includegraphics[width=0.98\linewidth]{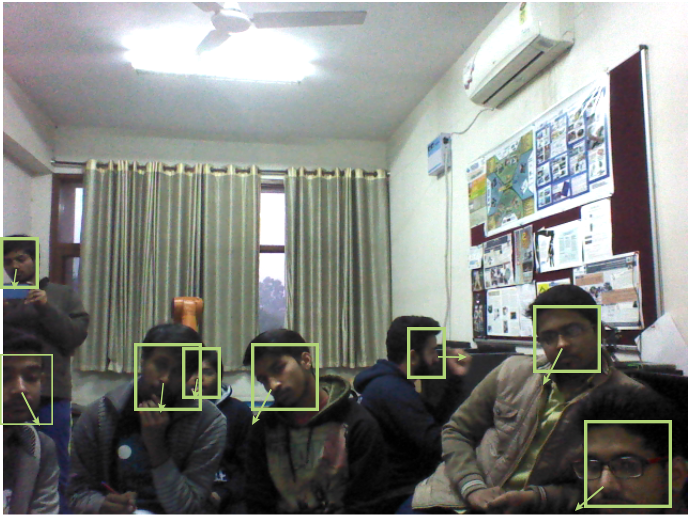}
        }
        \label{ques1}\hfill
    \subfloat[Image showing people evaluating the performance and filling the response sheet.]
        {
            \includegraphics[width=0.98\linewidth]{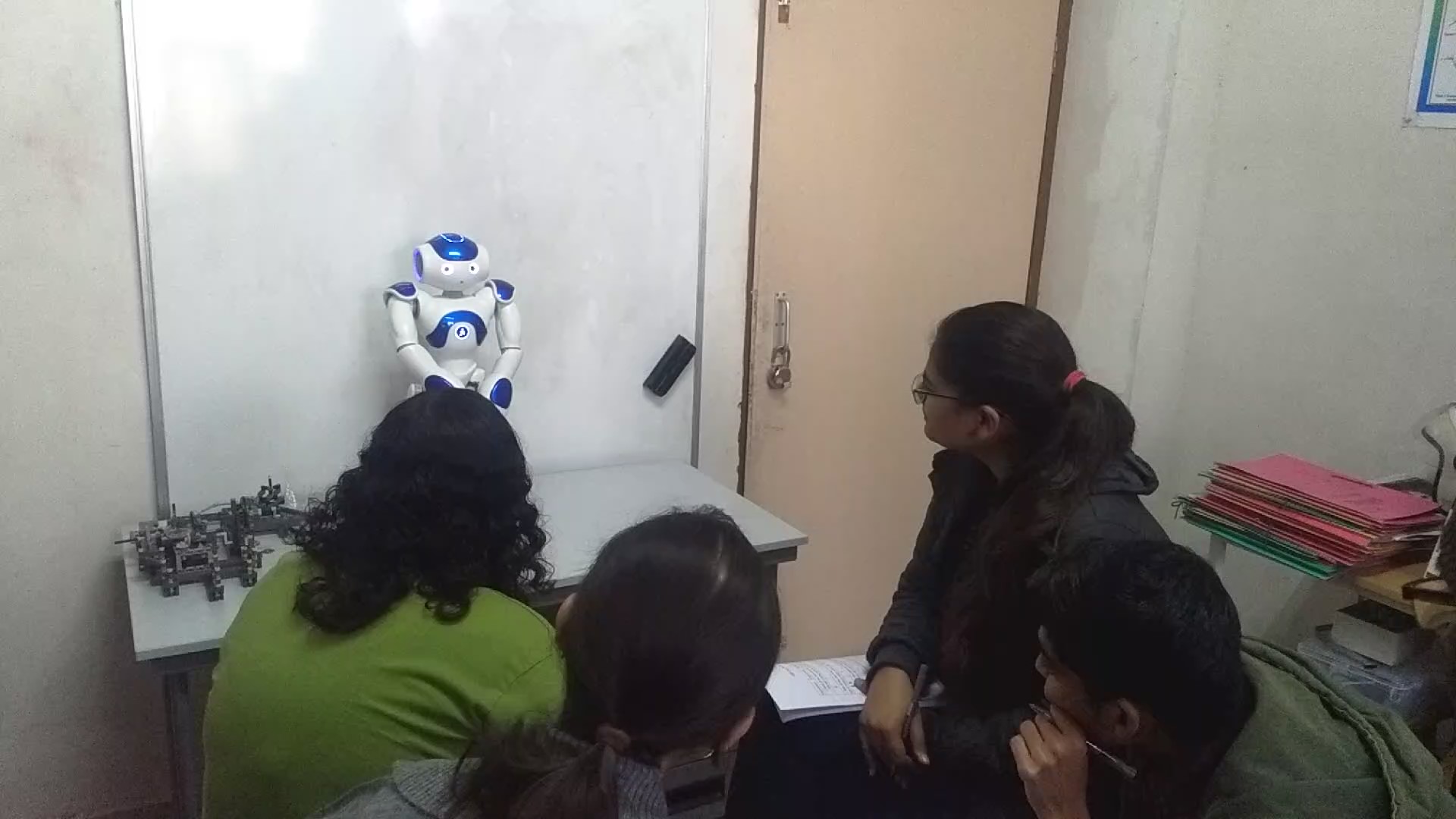}
        }
        \label{ques2}

      \caption[]{Images taken during the study.\protect \footnotemark }
      \label{exp}
\end{figure}
\footnotetext{Demo video of the experiments can be found at the attached link - https://youtu.be/Ws3G2M6aLto}
Research in the field of human-robot interaction (HRI) is gradually taking shape and a lot of work is going on to make the robots more sociable. There are many attributes to HRI like displaying emotional attributes, verbal communication attributes, visual scene understanding attributes and physical interaction attributes. To find an all encompassing scenario, we searched many possible scenarios where robots need all the above-mentioned attributes and in doing this we found out that social interaction using robots require verbal communication attribute to convey information, emotional and physical attributes to display intent and visual scene understanding attribute to assess the effect of conversation on the listener. The same has been explained by Gabbott et al. \cite{c11}, that a robot needs to understand and produce verbal and nonverbal signals in order to communicate with people and provide service. Thus the objective of this paper was decided to integrate all the above-mentioned attributes and utilize them to the fullest in HRI session and further investigate whether incorporating non-verbal gestures and modulation in speech patterns helped in increasing the acceptance rate by human participants. To assess the subject’s participation or degree of connection with the robot, the analysis of visual feedback is required and to perform non-verbal gestures robot needs physical attributes. This enables the conveyance of the speaker's intent to the audience and build a platform to test new and innovative methodologies that can be applied to improve limits of social communication. All of this is based on the hypothesis that non-verbal gestures and speech pattern play an important role in public speaking and incorporating them in a humanoid will lead to increased success rate of HRI sessions.

In 1980, Mehrabian et al. \cite{c1} experimentally showed that 55\% meaning of any message by people is generated by gestures. Another 38\% is derived from the speech pattern (tone, intonation, volume, pitch) and only 7\% from the said words. Mehrabian's results are only applicable when there is incongruity in the gesture and said word. R. Subramani\cite{c25} showed that 65\% of meaning during any given communication in Tirukkural(India) is conveyed via nonverbal communication. Kleinsmith et al. \cite{c12} also argued that gestures are an integral part of nonverbal communication. Despite the varying results of above mentioned researchers, we can easily deduce that nonverbal communication has an important role to play in our daily communication. Thus their work serves as the foundation for our hypothesis. To further investigate our hypothesis we tried to do the following-
 \begin{itemize}
     \item To create an end-to-end system that does the following - 
     \begin{itemize}
         \item maps speech patterns with body generated gestures as well as the rise and fall in pitch of the speaker.
         \item pays attention to engagement of the audience.
         \item adapts the speech pattern according to the audience engagement.
     \end{itemize} 
     \item To evaluate the acceptance of the system by the audience by conducting interaction survey as shown in Fig \ref{exp}.
 \end{itemize}
    
\begin{figure}[t]
      \centering
      \includegraphics[scale = 0.38]{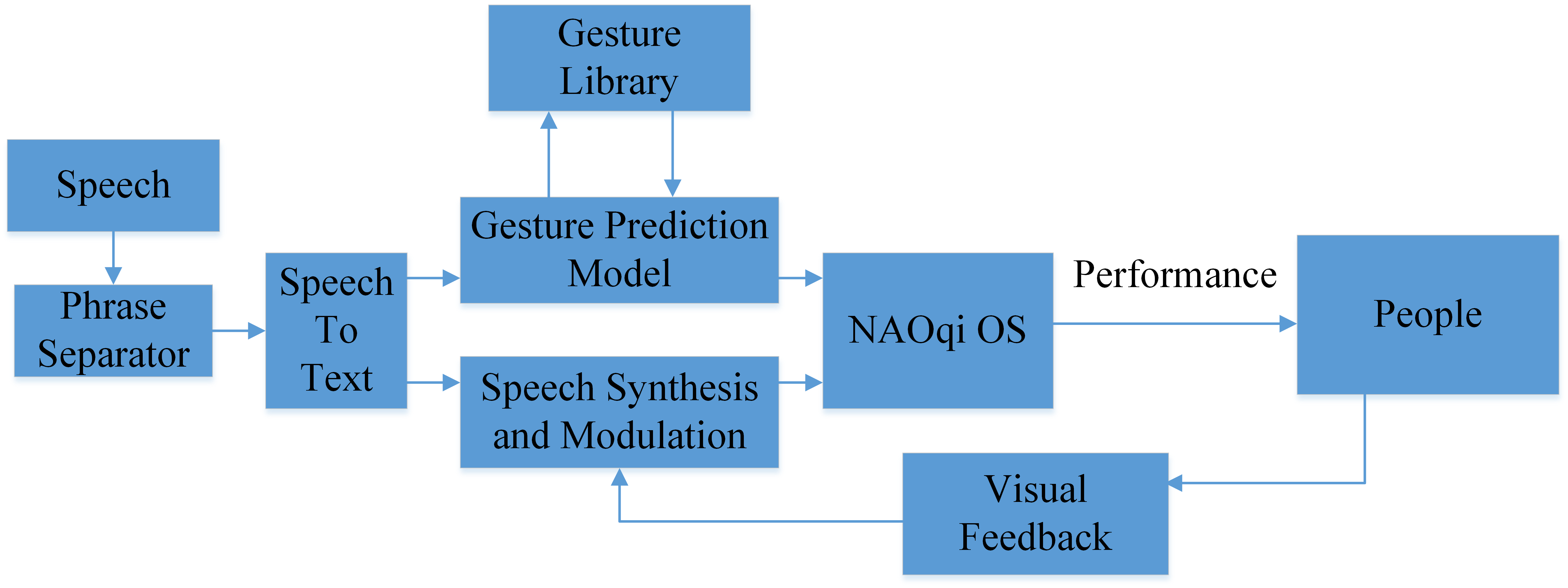}
      \caption{Flow diagram of our proposed method.}
      \label{flow}
\end{figure}

This would help us in deciding if any meaningful conversation is taking place or not, and with how much concentration the listener is listening to the speaker. As shown in Fig \ref{flow} the proposed method takes recorded audio as input and breaks it in phrases before converting into text. The text is then sent as an input to the learned model and it outputs the corresponding gesture from gesture library. The same text is also fed to speech synthesizer which produces the speech and modulates the speech based on the feedback received. The output of the model and the speech synthesizer is fed to NAOqi OS for performance. All the programming blocks are discussed in more details in later sections.  

Rest of the paper is organized as follows. Section II provides a discussion of previous work, section III provides insight into the creation of dataset, section IV discusses the training of the model and feedback system, section V describes the experimental setup, Section VI give the results of the experiment and section VIII concludes the paper with suggestion of some future areas of research in this particular domain.

\section{Related Work}
There have been several attempts to formalize the gesture synthesis process. Out of which the research by Junyun Tay and Manuela Veloso \cite{c26} has resulted in the most comprehensive gesture collection. They created gestures from key-frames, which they have permuted to create different gesture primitive. They have associated bag of words (BOW) for each gesture primitive and performed gesture primitive based on maximum overlap between BOW and input text. In their previous work \cite{c28}, they described their method for creating the key-frame collection. In their study, the authors trained a few high school kids to work on Choregraphe \cite{c29} and asked them to create two to three motions for each labels. Which we found out to be subjective and so we planned on creating our own key-frame dataset which is based on their real life usage. Meena et al. \cite{c3}, manually created the gesture library and defined limits on the scope for every gesture. For gesture performance, they created extension, key and retraction phases separately. For the speech synthesis, they extracted the text and then used a punctuator to identify utterance boundary. Ramachandran et al. \cite{c7}, mapped 5 gestures to 5 different emotions and used NAO as the mediator to attract attention and tempt children with autistic spectrum disorder (ASD) to participate in guessing game to instill emotion in them. The work by Meena et al. and Ramachandran et al. involved development of gestures, but neither of them showed the reason for gesture selection or the applicability of those gestures in selected scenario. 

People have also attempted to introduce expressibility in behaviour through the use of animations in place of robots like - Cassell et al. \cite{c27} created behaviour expression animation toolkit for creating animations with non-verbal gestures for on any input text. Gestures are performed based on hard wired rule set and for cases outside the rule set, default gesture is performed. Improving on their work Ng-Thow-Hing et al. \cite{c30} associated occurrence probability and expressivity parameter with gestures. This enhanced the co-occurrence of appropriate gestures and expression of emotions.   

HRI sessions to evaluate the efficiency of deployed systems have been attempted by many researchers. Meena et al. evaluated their model by mapping expectation and experience of participants through questionnaire. Ramachandran et al. evaluated their approach using two methods, firstly they asked the participants to appear in a pre-test and a post-test prepared on the contents of session to access the participant's learning within a single session. Secondly, they kept tracked the number of declined hints and auto hints and compared them across sessions to deduce if sessions are fruitful. In another similar approach, Ismail et al. \cite{c10} argued that eye contact plays an equally important role in understanding the quality of communication. That is why they proposed a method for detecting concentration level of child with ASD in its interaction with NAO. They performed gaze detection manually for the same purpose. 

Therefore, our proposed method uses gestures selected based on their usage density, which will be discussed in further detail in the following sections. It also uses eye gaze as a means of feedback as its applicability and effectiveness was demonstrated by Ismail et al. For the purpose of learning a model to generate relevant gestures, we were unable to find any relevant dataset. That is why, we created our own dataset. The steps involved in the creation of our dataset is discussed in the following section.  

\section{Creating the Dataset}

\subsection{Selecting Training Videos}

TED talks consists of public speeches by famous personalities all across the globe and span across different genres. Therefore, we selected popular TED talks on the basis of the views and amount of time camera was focused on the speaker. Selection of videos based upon the above-mentioned criteria was done manually and only 20 videos were selected.\\
From each of the videos 200 equally spaced frames were extracted. The frames which did not contain speaker were manually removed from the collection. In all of the frames, Convolutional Pose Machine (CPM) \cite{c4} is used to find out the location of the head and neck of the speaker. All these frames were then scaled and translated to align the speakers based on the extracted features.

\begin{figure}[t]
      \centering
      \includegraphics[scale=0.35]{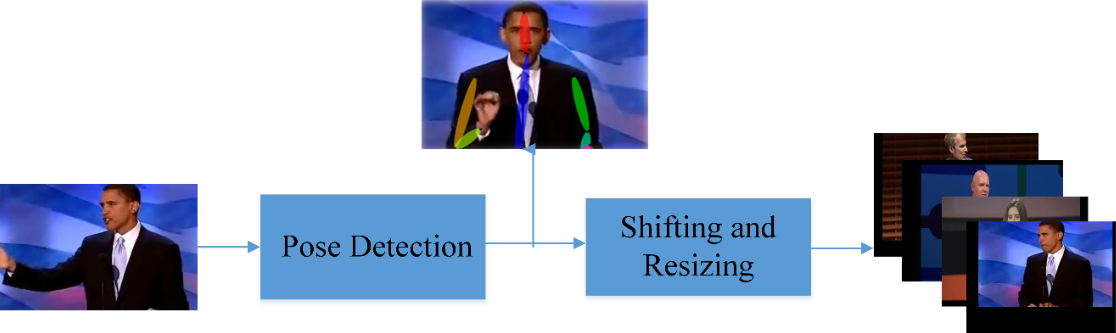}
      \caption{Flow diagram showing dataset homogenization.}
      \label{shift}
\end{figure}

\begin{equation}\label{eq1}
    ds_{i} = p_{i forehead} - p_{i Neck}
\end{equation}
\begin{equation}\label{eq2}
    I'_{i} = I_{i}* \frac{ds_{0}}{ds_{i}}
\end{equation}
\begin{equation}\label{eq3}
    p'_{i forehead} = shift(p_{i forehead},(p_{i forehead}-p_{0 forehead}))
\end{equation}

In Eq. \ref{eq1},~$ds_{i}$ refers to the head size of a person present in~$i^{th}$ image and~$p_{i xyz}$ refers to the location of the xyz point in~$i^{th}$ image. In Eq. \ref{eq2},~$I'_{i}$ refers to the final size of~$i^{th}$ image,~$I_{i}$ refers to initial size of~$i^{th}$ image and~$ds_{0}$ refers to the head size in reference image. In Eq. \ref{eq3},~$p'_{i forehead}$ refers to the head position of the person after image translation in~$i^{th}$ image,~$p_{i forehead}$ refers to head location of the person before translation in~$i^{th}$ image and function shift(a, b) shifts image \textit{a} by a distance \textit{b} . The result of resize and translation is shown in Fig \ref{shift}.
\begin{figure}
    \centering
    \subfloat[A hold A front]{\includegraphics[width=2.5cm, height=1.5cm ]{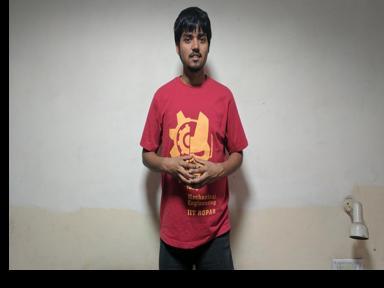}}
    \hfil
    \subfloat[AA front]{\includegraphics[width=2.5cm, height=1.5cm ]{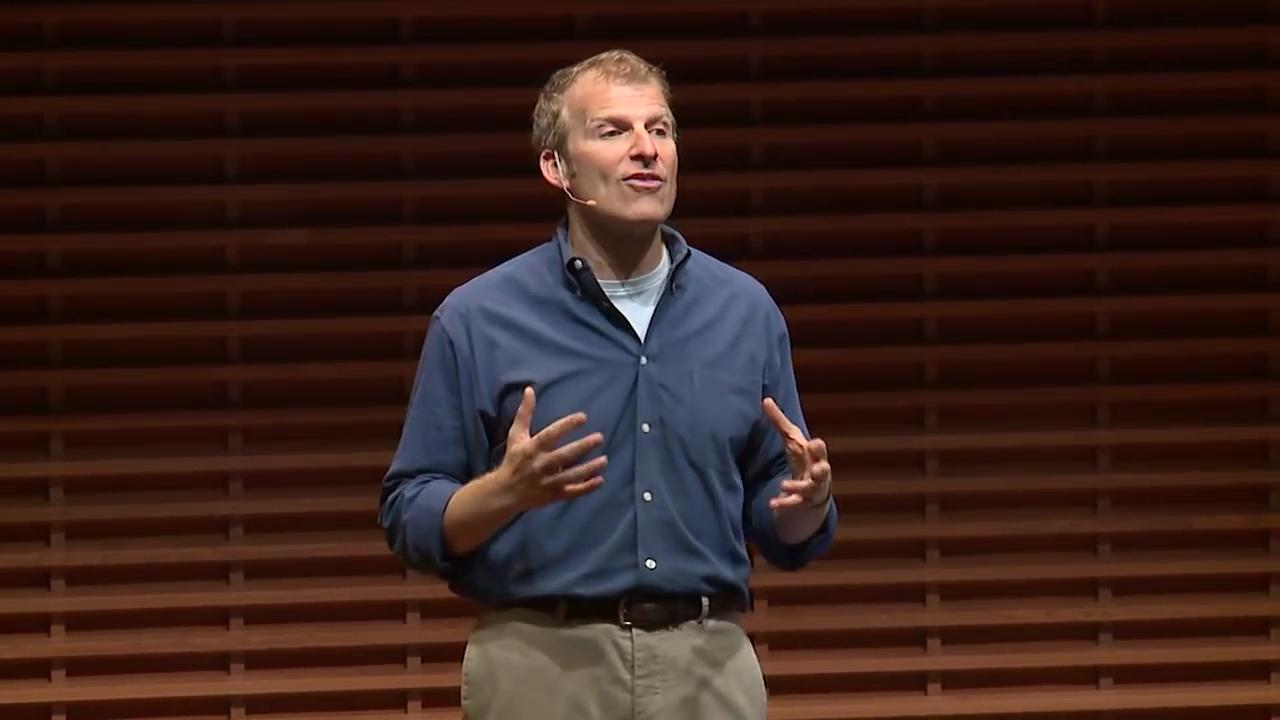}}
    \hfil
    \subfloat[RA Waits-T + LA front-T]{\includegraphics[width=2.5cm, height=1.5cm ]{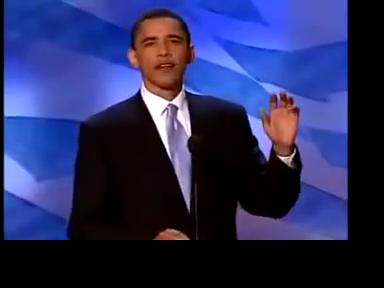}}\\
    \subfloat[LA side-T + RA side-T]{\includegraphics[width=2.5cm, height=1.5cm ]{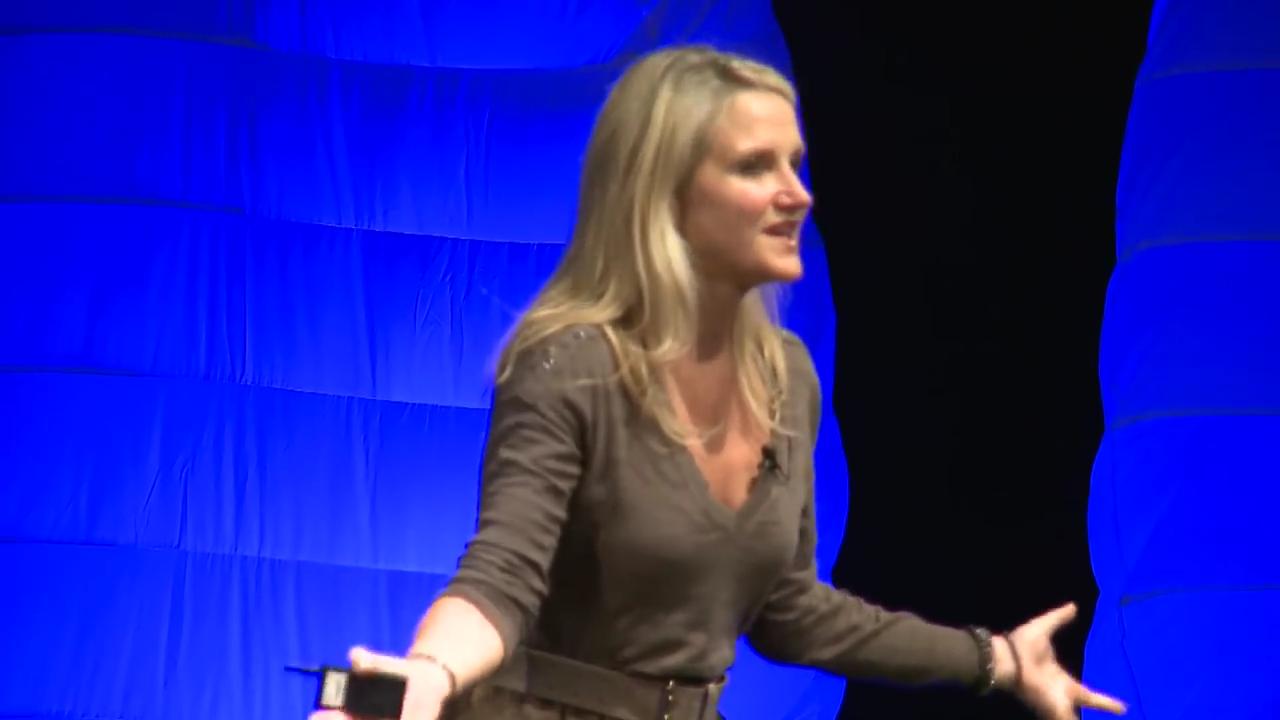}}
    \hfil
    \subfloat[LA front + RA front]{\includegraphics[width=2.5cm, height=1.5cm ]{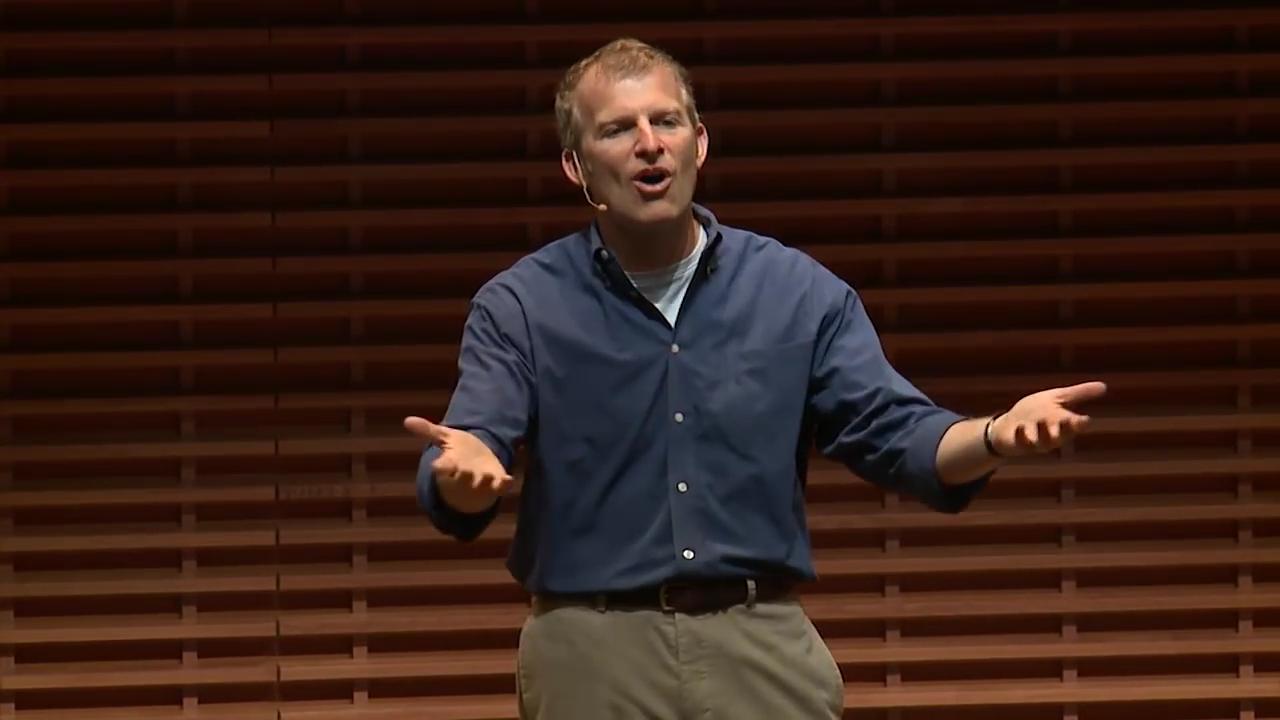}}
    \hfil
    \subfloat[LA side + RA front]{\includegraphics[width=2.5cm, height=1.5cm]{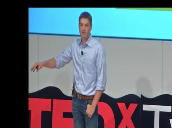}}\\
    \subfloat[LA Side + RH Waist-T]{\includegraphics[width=2.5cm, height=1.5cm ]{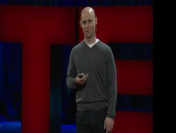}}
    \hfil
    \subfloat[LA Side + RA front-T]{\includegraphics[width=2.5cm, height=1.5cm ]{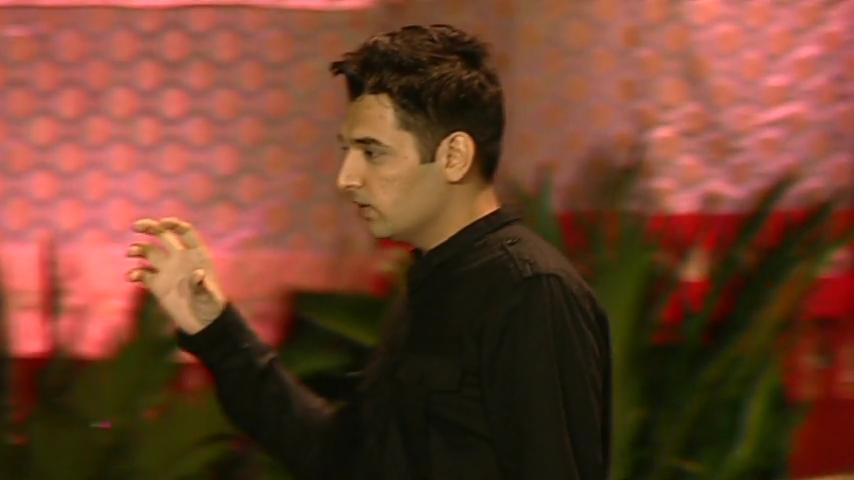}}
    \hfil
    \subfloat[LH pocket-T + RH pocket-T]{\includegraphics[width=2.5cm, height=1.5cm ]{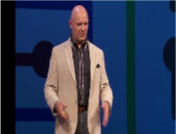}}
    
\caption{Frames nearest to the means in the 9 clusters named with BAP coding system \cite{c17, c18, c19, c20, c21, c22}.}
\label{cluster}
\end{figure}
%\begin{figure}[t]
%      \centering
%      \framebox{\parbox{3in}{\includegraphics[scale=0.52]{9cluster}}}
%      %\includegraphics[scale=1.0]{figurefile}
%      \caption{}
%      
%\end{figure}

\subsection{Determining Dominant Gestures}
For the purpose of identifying dominant gestures, CPM was used to identify the location of shoulder, elbow and wrist of both the arms. These coordinates were converted into relative distance vector from the neck. The set of vectors containing all the relative distances are then clustered using K-means algorithm \cite{c8}. The optimal value of number of clusters (k) and tolerance value~$(\epsilon)$ were chosen empirically to be 10 and 0.001 respectively. Out of the 10 clusters obtained from k-means, nine of them had gestures that conformed to the gesture nearest to the cluster center and last one was a collection of all the residual gestures. This implies that apart from 9 dominant gestures, there were no other high density gesture region in our 8 dimensional vector space. Center of these clusters are shown in Fig \ref{cluster} and the naming convention is adopted from Body Action and Posture (BAP) coding system \cite{c31}. For example - ``A hold A front" in 4(a) means that one hand is holding the other in front.

These cluster centers are significant for understanding our approach as they are not any randomly created gestures, but are the gestures used predominantly by the selected TED speakers in their speeches. Another advantage of the proposed approach is that the database creation process is fully automatic and requires minimal human input.

\subsection{Creating Gesture Library}
We created the gesture library using the cluster centers as the reference. Then using trial and error in joint jogging mode, we adjusted the joint angle parameters to the ones that result in similar body gestures on NAO humanoid as in the cluster centers. These full body joint angle configurations were saved as template files to create the gesture library.

\subsection{Audio Sampling}

\begin{figure}[t]
      \centering
      \includegraphics[scale=0.56]{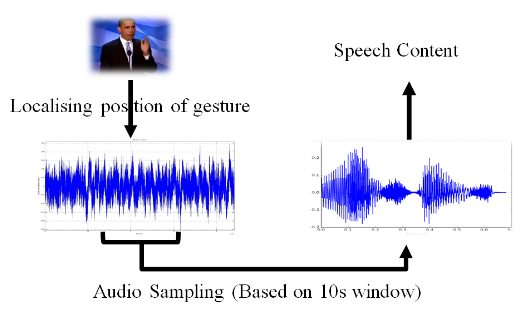}
      \caption{Flow diagram showing speech extraction.}
      \label{speech}
\end{figure}

In order to understand the context and to extract the meaning of the phrases, we needed to extract the content of the corresponding speech. For this purpose, an audio of length 10s was extracted centered around the saved gesture frames and later a dynamic phrase level speech extraction was employed as shown in Fig \ref{speech}.\\
To extract content from the speeches, we used the Sphinx library \cite{c23}. During the translation phase, it was found out that sphinx when used without dictionary and grammar support is biased towards native American speakers and we also found out that accuracy decreased in case of female speakers.

\subsubsection{Sampling Issues}
During our initial trials, it was found that a speech segment of 10s did not represent the contents properly for which speaker was performing gestures. In our further investigation, we found out that, in normal scenario, every gesture is performed for a single phrase. So, we extracted the central phrase from the 10s audio segment.

\subsubsection{Phrase Boundary Marking}
We computed finite interval fast Fourier transform \cite{c15} to capture the ~$0^{th}$ formant (first frequency with the highest intensity). After that, we applied a low pass filter to remove noise and convolved the signal with 1-D Gaussian filter to smoothen out the residual noise. In the perceived output, the regions with low energy for longer duration were considered as phrase boundary. This can also be seen in Fig 6a.

\subsubsection{Phrase Boundary Separation}
\begin{figure}[t]
      \centering
      \includegraphics[scale=0.65]{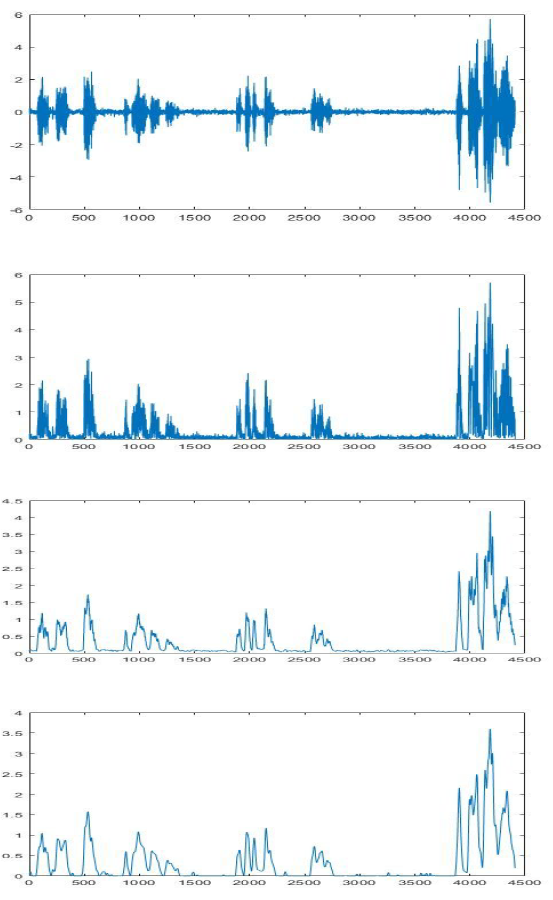}
      \caption{Plot of a) ~$F_{0}$ b) Absolute value of ~$F_{0}$ c) After max-pooling and average pooling d) After applying threshold.}
      \label{audio}
\end{figure}

As we needed to extract the phrase boundary information, we took the absolute value of the ~$F_{0}$ obtained in the previous step and performed max-pooling on them, and after that we performed average pooling to filter out the sharp irregularities. Finally, a threshold at 30\% of maximum intensity was applied to remove the last remaining noise. From the time period where intensity reaches 0 for 0.2s, glottal opening and closing are calculated. The output of each stages are shown in Fig \ref{audio}. This created an issue of over-classification of fillers (i.e-oh, um...) and conjunctions (i.e and, but...) as phrases. To overcome the problem of conjunction separation we merged the small phrases with nearby phrases whose duration was less than 100 bins. This helped in avoiding dangling fillers and conjunctions. The parameters used in phrase separation is subjected to change based on the pool from which speaker is selected. 
   
Summary of the information extracted from the audio sampling are listed below-
\begin{itemize}
    \item time stamps corresponding to glottal opening and closing.
    \item pause duration between phrases.
    \item duration of individual phrases.
\end{itemize}

\section{Model And Feedback System}

In this section, the training model selected for learning the map between speech and gestures in the gesture library is discussed in detail. Discussion on why and how the feedback system is created is also presented in this section.

\subsection{Training the Model} Due to small number of training samples available left after filtering, we could not opt for training method which involved neural networks. Due to that specific reason, the mapping between gesture and the associated speech is carried out using a supervised learning algorithm called \textit{Decision Tree} \cite{c14}. It is one of the most frequently used algorithm in operations research and in machine learning as well. We used group of differently trained decision trees also known as a random forest. A Random forest is a group of decision trees trained on different attribute sets or instance sets.\\ 
In this work, random forest was trained on training instances having an average of 12 words per gesture as attributes. A total of 500 trees were trained with feature bagging. This random forest outputs a name --- one of the nine gestures which is best suitable for the given input speech.

\subsection{Preparation of Visual Feedback System}
A feedback system is such a system in which an error or a portion of the output signal is sent back as input system by closing the loop. This error output signal is then used to modify input signal so that the error in output signal reduces.

Lemaignan et al. \cite{c2} proposed an innovative way to assess with-me-ness in real time, which served as the inspiration behind our feedback system. In current case, the images captured using the camera mounted on the head of Nao are used to compute the gaze vector of people sitting in the audience. To compute this, Openface library \cite{c16} was used. For the calculation purpose, we assumed the distance along the z-axis to be 5m. If the endpoint of gaze vector on Nao's plane lied within a radius of 2m of Nao's center then we considered that person was attentive else the person was not attentive. If the percentage of the attentive audience dropped below 50\% then the mean pitch and mean volume of Nao was raised by 10\% every 15s. If the case was other way around then pitch and volume were reduced at a rate of 10\% per 15s till they reached the base values.   

\begin{figure}[t]
      \centering
      \includegraphics[scale=0.5]{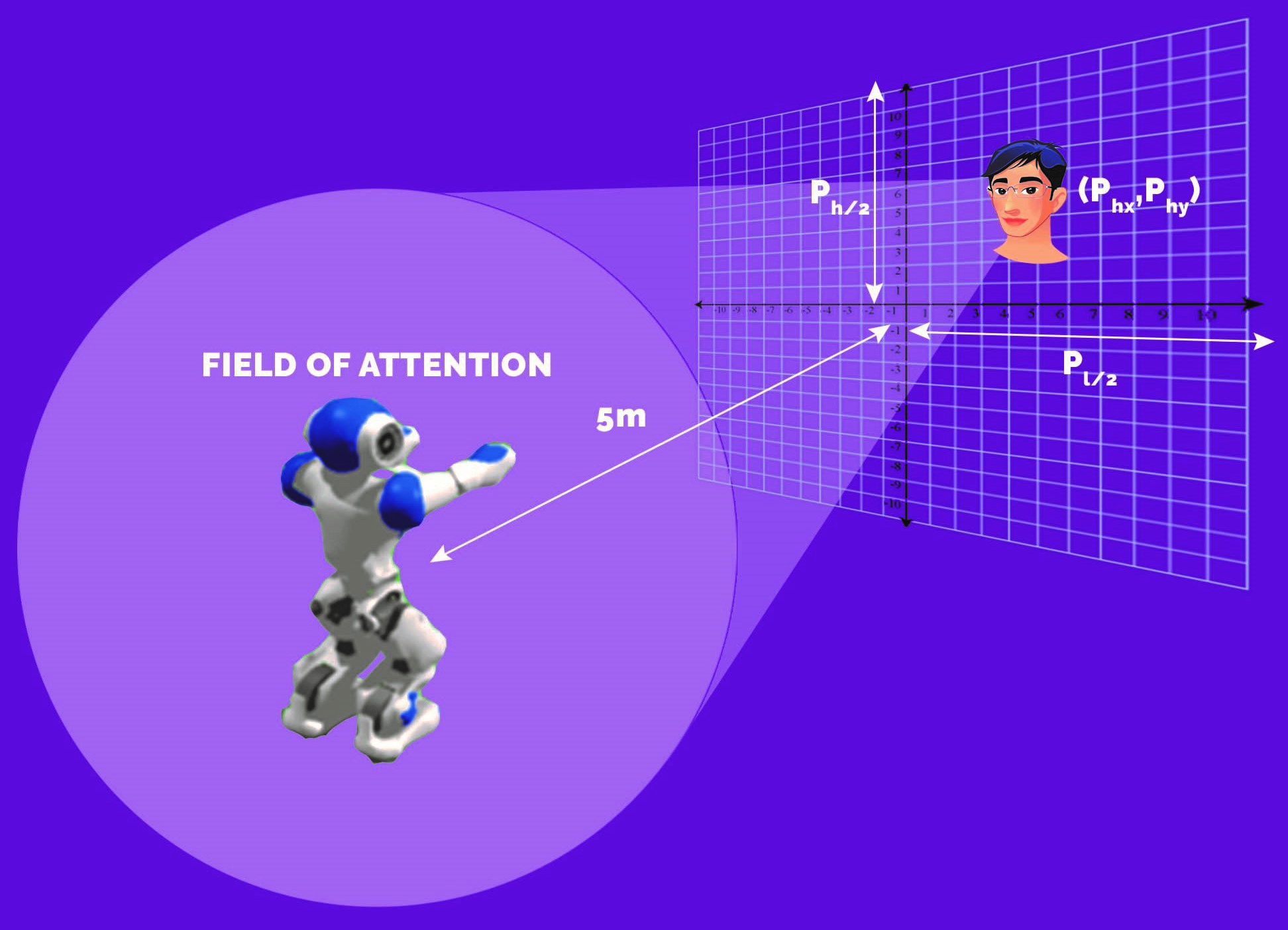}
      \caption{Diagram showing the field of attention and other parameters associated in attention tracking.}
      \label{fov}
\end{figure}

We used the following equation to compute attentiveness. 
\begin{equation}\label{eq4}
    i_{bx} = 5*\tan{\frac{60.97}{2}},
\end{equation}
\begin{equation}\label{eq5}
    i_{hx} = \frac{p_{hx}*i_{bx}}{p_{\frac{l}{2}}} .  
\end{equation}
In Eq \ref{eq4} ~$i_{bx}$ is the distance between a point in real world that lies on the edge of the image and another point in real world that lies at the center of the image, both at a perpendicular distance of 5m from NAO's head-cam as shown in Fig \ref{fov}. The camera width angle along x-axis is 60.97\degree. In Eq \ref{eq5} ~$i_{hx}$ represents the x-coordinate of a person's head in the real plane located at a distance of 5m from NAO, ~$p_{hx}$ represents the x-coordinate of a person's in image plane and ~$p_{\frac{l}{2}}$ represents half of image width. Eq \ref{eq5} is derived from the result of basic proportionality theorem applied on similar triangles.

Using the above calculations for y-axis,
\begin{equation}\label{eq6}
    i_{by} = 5*\tan{\frac{47.64}{2}}
\end{equation}
\begin{equation}\label{eq7}
    i_{hy} = \frac{p_{hy}*i_{by}}{p_{\frac{l}{2}}}  
\end{equation}
If the gaze vector V is [~$d_{x}, d_{y}, d_{z}$] then...
The end point of gaze vector on Nao's plane will be given by
\begin{equation}\label{eq8}
    g_{end} = [\frac{d_{x}}{d_{z}}*5, \frac{d_{x}}{d_{z}}*5, 5]
\end{equation}
Corrected coordinates~$\alpha$ \&~$\beta$ were calculated after shifting back the head position from image center 
\begin{equation}\label{eq9}
    [\alpha, \beta] = [\frac{d_{x}}{d_{z}}*5 - i_{hx}, \frac{d_{x}}{d_{z}}*5 - i_{hy}]
\end{equation}
A person is considered attentive if
\begin{equation}\label{eq10}
    {\alpha}^2+{\beta}^2 < 4
\end{equation}

A pitch modulation algorithm was implemented drawing inspiration from Langarani et al \cite{c6}. A linear mapping between mean pitch and the attention level in place of GMM mapping was implemented. If the number of people whose attention wavers increased more than 50\%, then the feedback response/speech modulation would start.

\section{Experimental Design}

\begin{figure*}
    \centering
    \subfloat[``Special operation force's troops"]
        {
            \includegraphics[width=0.18\linewidth, height=3in]{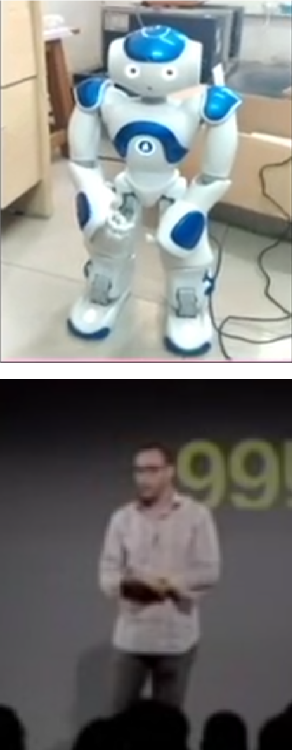}
        }
        \label{pn1}
    \subfloat[``One of the pilots up above"]
        {
            \includegraphics[width=0.18\linewidth, height=3in]{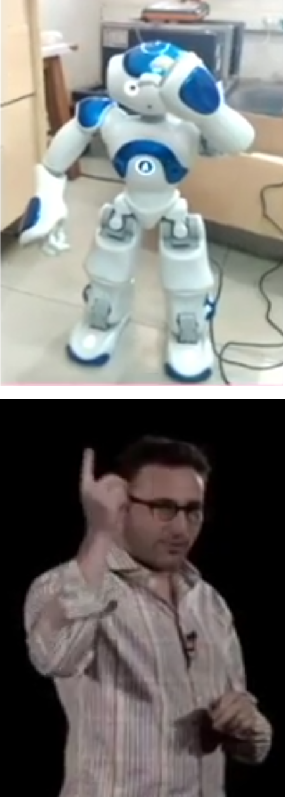}
        }
        \label{pn2}
    \subfloat[``And, yes he stands like this"]
        {
            \includegraphics[width=0.18\linewidth, height=3in]{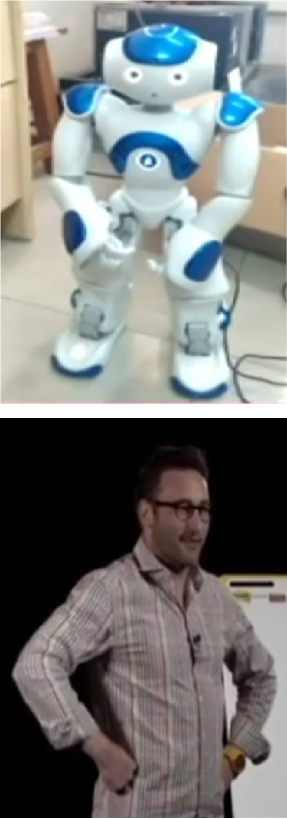}
        }
        \label{pn3}
    \subfloat[``Tells his wing man to hang out"]
        {
            \includegraphics[width=0.18\linewidth, height=3in]{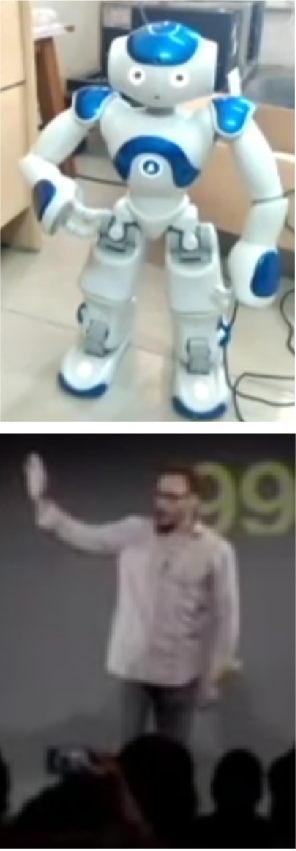}
        }
        \label{pn4}
    \subfloat[``The plane is getting thrashed about"]
        {
            \includegraphics[width=0.18\linewidth, height=3in]{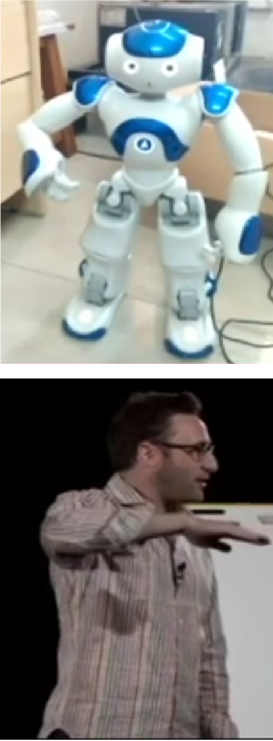}
        }
        \label{pn5}
    \captionsetup{justification=centering}
    \caption{ Gesture exhibited by speaker \cite{c32} and NAO for captioned context. Complete comparison video can be found at \url{https://www.youtube.com/watch?v=ZBCTmD4xiaA}}
    \label{comp}
\end{figure*}

To understand the extent of improvements provided by the proposed methodology we performed a comparative study as shown in Fig \ref{comp} and an experimental setup was designed to evaluate the performance, as seen by the humans with whom the robot was interacting. For this, behaviours from different stages of its development and a few questions for the onlookers is prepared. The questions and scenarios are explained below. Behaviours in the survey consisted of the following scenarios.

\begin{itemize}
    \item \textbf{Speech without any adaptation or gesture:} In this scenario, NAO was given only the text input to speak. This experiment demonstrates the effectiveness of communication performed by a humanoid prior to this work, which also serves as the baseline for our evaluation.
    \item \textbf{Speech with gesture of 10 seconds period without any adaptation:} For this, NAO was given speeches of 10 seconds window. It had to predict what kind of gesture it needed to perform for the given speech and then using the gesture library, as mentioned in section III, it performed the gesture along with the speech.
    \item \textbf{Speech with phrase level gesture period without any adaptation:} For this, phrases were detected and then fed to NAO to perform phrase level gesture prediction. These phrase level gestures were performed along with the phrases for which these were predicted.
    \item \textbf{Phrase level speech adaptation with phrase level gesture period:} For this, we included speech modulation along with the phrase level gesture prediction. In order to achieve this, we mapped the humanoid's pitch and intonation pattern to the target speaker’s pitch and intonation pattern.
\end{itemize}

A user survey was conducted across the following four questions to obtain participant's view on the proposed system.

\begin{itemize}
    \item \textbf{Rank all the four behaviours mentioned above based upon their similarity to humans:} In this, the users were supposed to rank all the behaviors relative to each other.
    \item \textbf{Rate all the four behaviours mentioned above based upon their similarity to human:} In this, the users were supposed to rate all the behavior experiments on an absolute scale of 1-10.
    \item \textbf{Rate all the four behaviours mentioned above based upon their gesture to speech synchronization:} In this, the users were supposed to rate how fluid or in sync were those behavior experiments on an absolute scale of 1-10.
    \item \textbf{Rate all the four behaviours mentioned above based upon how improvised the generated gestures are?:} In this, the users were supposed to rate meaningfulness of the gesture performed in all the behavior experiments on an absolute scale of 1-10.
\end{itemize}

\section{Results}
\begin{figure*}[t]
    \centering
    \subfloat[Average of comparative ranks assigned by attendees to different behaviors based on their human likeness.]
        {
            \includegraphics[width=0.48\linewidth]{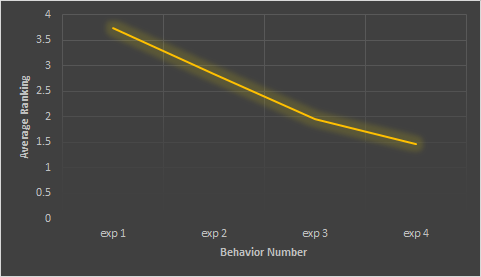}
        }
        \label{qu1}\hfill
    \subfloat[Average of the scores (out of 10) assigned by attendees to different behaviors based on their human likeness.]
        {
            \includegraphics[width=0.48\linewidth]{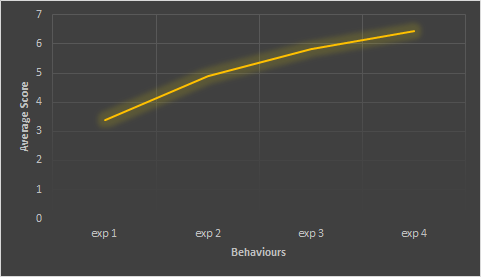}
        }
        \label{qu2}\\
    \subfloat[Average of the scores (out of 10) assigned by attendees to different behaviors based on their gesture to speech synchronization.]
        {
            \includegraphics[width=0.48\linewidth]{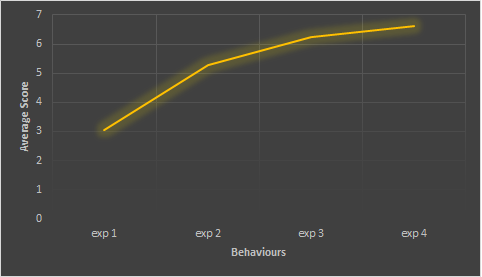}
        }
        \label{qu3}\hfill
    \subfloat[Average of the scores (out of 10) assigned by attendees to different behaviors based on self-improvisation of the generated gestures.]
        {
            \includegraphics[width=0.48\linewidth]{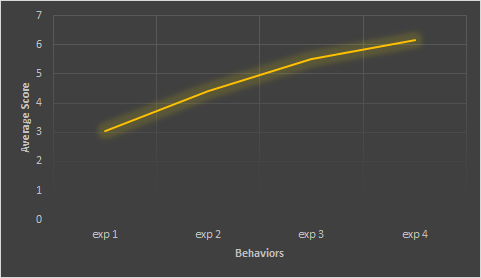}
        }
        \label{qu4}
    \caption{Participant's response to different questions.}
    \label{question}
\end{figure*}

The interaction study was conducted over multiple sessions consisting of ~$4\sim6$ participants each. All the 4 behaviour experiments as mentioned in section V were shown to them at random order. Each behaviour experiment covered an extract of 10 minutes from famous speeches and lasted for $7\sim10$ minutes based on the model employed. Content covered by the robot varied across the sessions, but were kept fixed across all the behaviours in a session. Only basic explanation about the questions were provided to the participants. In total 24 people participated. Among them 12.5\% were females and rest were males. Equal number of graduates and undergraduates turned up for this survey. Out of which 16.67\% were from Civil department, 16.67\% from Electrical Department, 29.17\% from Computer Science Department and 37.5\% from Mechanical Department.

In the experiments, we observed a rise in acceptance rate by the audience. We used ANOVA method \cite{c24} to analyze inter-experiment variance and to validate the significance of the addition of each new feature on the acceptance by the audience. For the computation of ANOVA, significance level ~$\alpha=0.05$ was used. In all the cases we obtained the probability of getting result in accordance with the null hypothesis ~$p-value<\alpha$. The obtained results are discussed in detail below.

\begin{itemize}
    \item \textbf{Rank all the four behaviours based on their similarity to human being}
        
        %\begin{table}[h]
        %    \caption{ANOVA results for Question 1}
        %    \label{q1}
        %    \begin{center}
        %        \begin{tabular}{|c|c|c|c|c|c|c|}
        %            \hline
        %            Variance Source & SS & df & MS & F & P-value & F crit\\
        %            \hline
        %            Between Groups & 73.25 & 3 & 24.42 & 48.05 & 8.9E-19 & 2.70\\
        %            \hline
        %            Within Groups & 46.75 & 92 & 0.508 & & &\\
        %            \hline
        %        \end{tabular}
        %    \end{center}
        %\end{table}
        
        The decreasing trend in the graph shown in Fig 9(a) represents how audience ranked the  system more human-like than the performance given by Nao's inbuilt speech synthesizer with autonomous life ``on". The scores obtained for behaviour ~$1\sim4$ are (min - 4, avg - 3.75, var - 0.45), (min - 4, avg - 2.83, var - 0.66), (min - 3, avg - 1.95, var - 0.39) and (min - 3, avg - 1.45, var - 0.57). The p-value obtained from one-way ANOVA was ${8.9e}^{-19}$.
        
    \item \textbf{Rate all the four behaviours based on their similarity to human being}
        %\begin{table}[h]
        %    \caption{ANOVA results for Question 2}
        %    \label{q2}
        %    \begin{center}
        %        \begin{tabular}{|c|c|c|c|c|c|c|}
        %            \hline
        %            Variance Source & SS & df & MS & F & P-value & F crit\\
        %            \hline
        %            Between Groups & 129.21 & 3 & 43.07 & 13.91 & 1.48E-07 & 2.70\\
        %            \hline
        %            Within Groups & 284.75 & 92 & 3.09 & & &\\
        %            \hline
        %        \end{tabular}
        %    \end{center}
        %\end{table}
        
        The increasing trend in the graph shown in Fig 9(b) represents how audience rated our current system as compared to robot speeches as seen in movies. The scores obtained for behaviour ~$1\sim4$ are (min - 0, avg - 3.37, var - 5.11), (min - 1, avg - 4.91, var - 2.94), (min - 2, avg - 5.83, var - 2.23) and (min - 3, avg - 6.45, var - 2.08). The p-value obtained from one-way ANOVA was ${1.48e}^{-07}$.
    \item \textbf{Rate all the four behaviours based on their gesture to speech synchronization}
        %\begin{table}[h]
        %    \caption{ANOVA results for Question 3}
        %    \label{q3}
        %    \begin{center}
        %        \begin{tabular}{|c|c|c|c|c|c|c|}
        %            \hline
        %            Variance Source & SS & df & MS & F & P-value & F crit\\
        %            \hline
        %            Between Groups & 186.2 & 3 & 62.06 & 21.14 & 1.67E-10 & 2.70\\
        %            \hline
        %            Within Groups & 270.04 & 92 & 2.93 & & &\\
        %            \hline
        %        \end{tabular}
        %    \end{center}
        %\end{table}
        
        The increasing trend in the graph shown in Fig 9(c) represents how audience rated our current system's synchronization between speech initiation and the beginning of gesture. This also includes whether the amount of pauses were accurate or not.  The scores obtained for behaviour ~$1\sim4$ are (min - 0, avg - 3.04, var - 5.59), (min - 2, avg - 5.29, var - 2.04), (min - 4, avg - 6.25, var - 2.02) and (min - 4, avg - 6.62, var - 2.15). The p-value obtained from one-way ANOVA was ${1.67e}^{-10}$.
    \item \textbf{Rate all the four behaviours based on how improvised the generated gestures are}
        %\begin{table}[h]
        %    \caption{ANOVA results for Question 4}
        %    \label{q4}
        %   \begin{center}
        %        \begin{tabular}{|c|c|c|c|c|c|c|}
        %            \hline
        %            Variance Source & SS & df & MS & F & P-value & F crit\\
        %            \hline
        %            Between Groups & 134.3 & 3 & 44.76 & 13.6 & 1.97E-07 & 2.70\\
        %            \hline
        %            Within Groups & 302.1 & 92 & 3.28 & & &\\
        %            \hline
        %        \end{tabular}
        %    \end{center}
        %\end{table}
        
        The increasing trend in the graph shown in 9(d) represents how audience rated our current system's success in conveying meaning with the help of gestures and how well those gestures could be correlated with the spoken words. The scores obtained for behaviour ~$1\sim4$ are (min - 0, avg - 3.04, var - 5.43), (min - 1, avg - 4.41, var - 2.94), (min - 2, avg - 5.5, var - 2.08) and (min - 3, avg - 6.16, var - 2.66). The p-value obtained from one-way ANOVA was ${1.97e}^{-07}$.
\end{itemize}

From the p-value obtained from the surveys we can draw the conclusion that our proposed method is statistically significant and using phrasal gestures and voice modulation improves performance over currently employed time bound gesture generation.

\subsection{Evaluation of the Feedback System}

\begin{table}[h]
    \caption{Mean value of attentiveness during different experiments}
    \label{attention}
    \begin{center}
        \begin{tabular}{|c|c|c|c|c|}
        \hline
         & Exp 1 & Exp 2 & Exp 3 & Exp 4\\
        \hline
        Attention Value & 73.64\% & 73.21\% & 74.02\% & 73.11\% \\
        \hline
        \end{tabular}
    \end{center}
    \end{table}

As seen in the table \ref{attention}, mean attentiveness never dropped below 50\% so we were unable to validate the improvement due to pitch and volume modulation. We argue that a potential reason for this anomaly might be the anticipation towards Nao humanoid among the audience. Because of that, we conducted the survey twice but the results were similar.

\section{Conclusion}
This paper end-to-end approach that enables humanoid robots to produce human like interactions in a social context. The core system consisted of three parts, which are the dataset creation part, the model generation part and the feedback part. The data set is created from real life speeches of TED speakers. The model for speech-gesture mapping is based random forest. The feedback uses attention tracking, Openface library had been used. 

Another key contribution of this paper is that it experimentally shows that translating human non-verbal gesture to HRI sessions improve the acceptance rate of robotic interaction by participants. We have demonstrated the performance of our system during those experiments with Nao humanoid. The approval of attendees clearly show the improvement made in the direction of conveying meaning as compared to standard audio speeches. We will be making the project code and gesture template available on Github.

In future work, we plan to extend the abilities of the robot towards learning diverse gestures using auto encoders and reinforcement learning as approached by Qureshi et al. [5]. Also making it capable of generating gestures in completely unseen environments while being contextually aware. To improve the social performance we plan to implement multi-modal deep reinforcement learning as already shown by them.





%\bibliographystyle{IEEEtran}

\end{document}